%% file: acl_latex.tex
\pdfoutput=1
\documentclass[11pt]{article}
\usepackage[preprint]{acl}

\usepackage{times}
\usepackage{latexsym}

\usepackage[T1]{fontenc}
\usepackage[utf8]{inputenc}

\usepackage{microtype}

\usepackage{inconsolata}

\usepackage{graphicx}
\usepackage{amsmath}
\usepackage{booktabs}
\usepackage[table]{xcolor}
\usepackage{caption} % figure subcaption위해서 추가
\usepackage{subcaption} % figure subcaption위해서 추가
\definecolor{lightblue}{RGB}{235, 245, 255}  % 연한 파란색 정의
\usepackage[most]{tcolorbox}
\usepackage{multirow}

\usepackage{hyperref}
\hypersetup{
    colorlinks=true,
    linkcolor=blue,
    filecolor=magenta,      
    urlcolor=cyan,
}

\usepackage{cuted}

\usepackage{tcolorbox}
\tcbuselibrary{breakable}

\newtcolorbox{promptbox}{
  breakable,
  colback=gray!5,
  colframe=gray!60,
  boxrule=0.5pt,
  arc=2pt,
  left=6pt,
  right=6pt,
  top=6pt,
  bottom=6pt
}

\title{Thinking-KT: A Training-Free Large Reasoning Model-based\\Knowledge Tracing Framework for Unified Prediction and Prescription}

\author{Unggi Lee\textsuperscript{1}, Joo Young Kim\textsuperscript{2}, Ran Ju\textsuperscript{2}, Minyoung Jung\textsuperscript{2$\dagger$}, Jeyeon Eo\textsuperscript{3$\dagger$} \\
  \textsuperscript{1}Chosun University, Gwangju, Republic of Korea \\
  \textsuperscript{2}Neudive Inc, Daegu, Republic of Korea \\
  \textsuperscript{3}Independent Researcher, Seoul, Republic of Korea
  \\\texttt{codingchild@korea.ac.kr}, \texttt{jyjoshk@kitech.re.kr}, \texttt{ran\_ju@neudive.com}\\ \texttt{minyoung@kbri.re.kr}, \texttt{jeyeon.yona.eo@gmail.com}
  }

\begin{document}
\maketitle

\input{0_abstract}
\input{1_intro}

\input{3_method}

\input{4_result}

\input{5_ablation}
\input{6_analysis}

\input{2_related_work}

\input{7_conclusion}

% \bibliographystyle{acl_natbib}
% \bibliography{custom}

\newpage

\clearpage 
\appendix

\input{8_appendix}

\end{document}

%% file: 0_abstract.tex
\begin{abstract}
Knowledge Tracing (KT) aims to estimate a learner’s evolving mastery based on interaction histories. Recent studies have explored Large Language Models (LLMs) for KT via autoregressive nature, but such approaches typically require fine-tuning and exhibit unstable or near-random performance. Moreover, prior KT systems primarily focus on prediction and rely on multi-stage pipelines for feedback and recommendation, resulting in increased system complexity and resources. To address this gap, we propose \textit{Thinking-KT}, a training-free KT framework that incorporates Test-Time Scaling (TTS), enabling even small LLMs (1–2B parameters) to achieve competitive KT performance. Moreover, in this framework, a small LLM can jointly perform KT prediction, personalized feedback generation, and learning recommendation in a unified output without degrading prediction accuracy. Beyond performance, we present the systematic analysis of reasoning traces in KT. Our results demonstrate that TTS is a critical yet underexplored factor in LLM-based KT, and that small LLMs can serve as unified ITS engines. Our code is available at Our code is available at \url{https://anonymous.4open.science/r/lokt_thinking_anonymous-2A16}
\end{abstract}

%% file: 1_intro.tex
\begin{figure}
    \centering
    \includegraphics[width=1\linewidth]{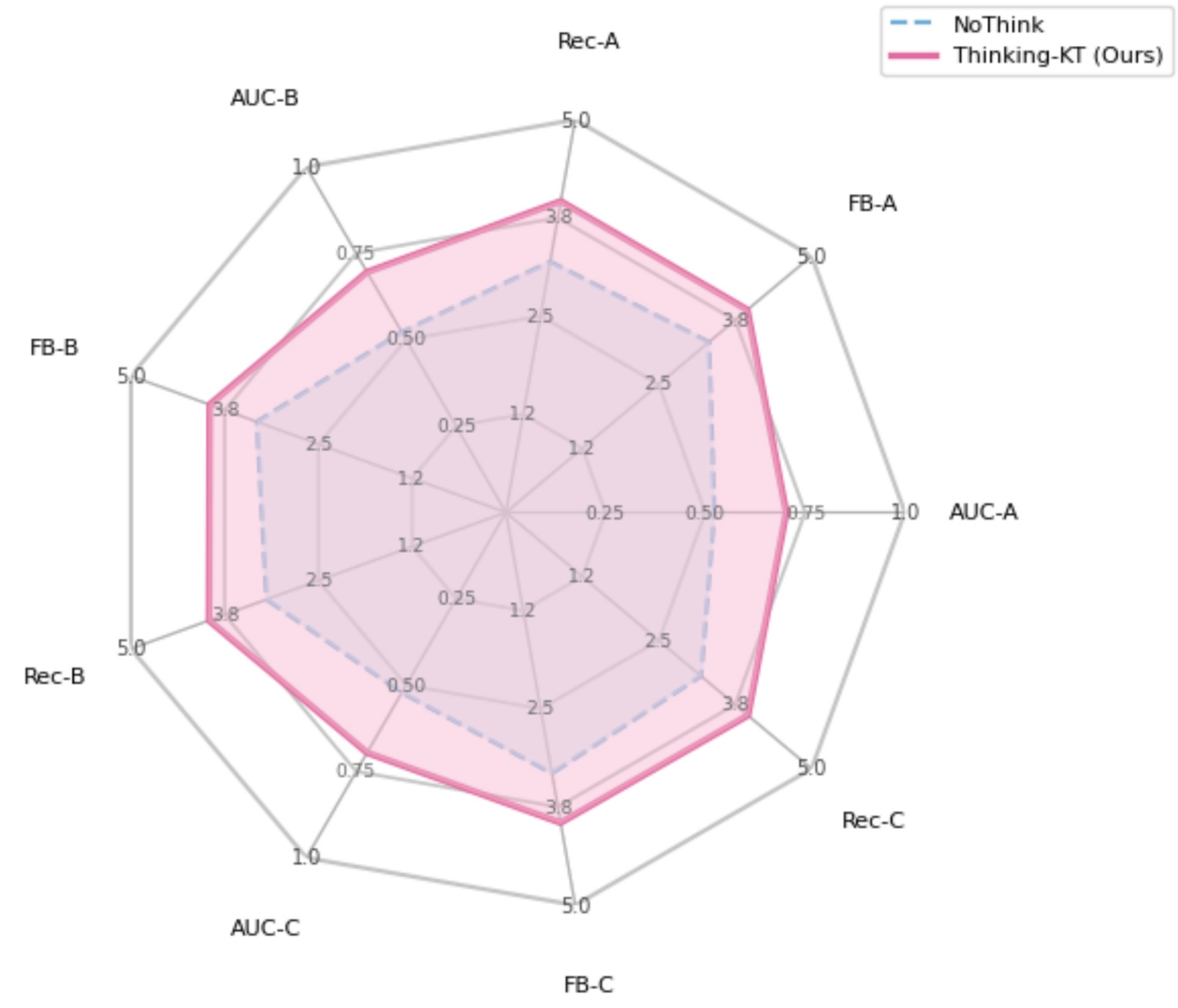}
    \caption{Comparing No-Think and \textit{Thinking-KT} across KT performance and pedagogical quality of feedback and recommendations. TTS consistently improves both prediction accuracy and pedagogical quality without training.}
    \label{fig:placeholder}
\end{figure}

\section{Introduction}

\begin{figure*}
    \centering
    \includegraphics[width=1\linewidth]{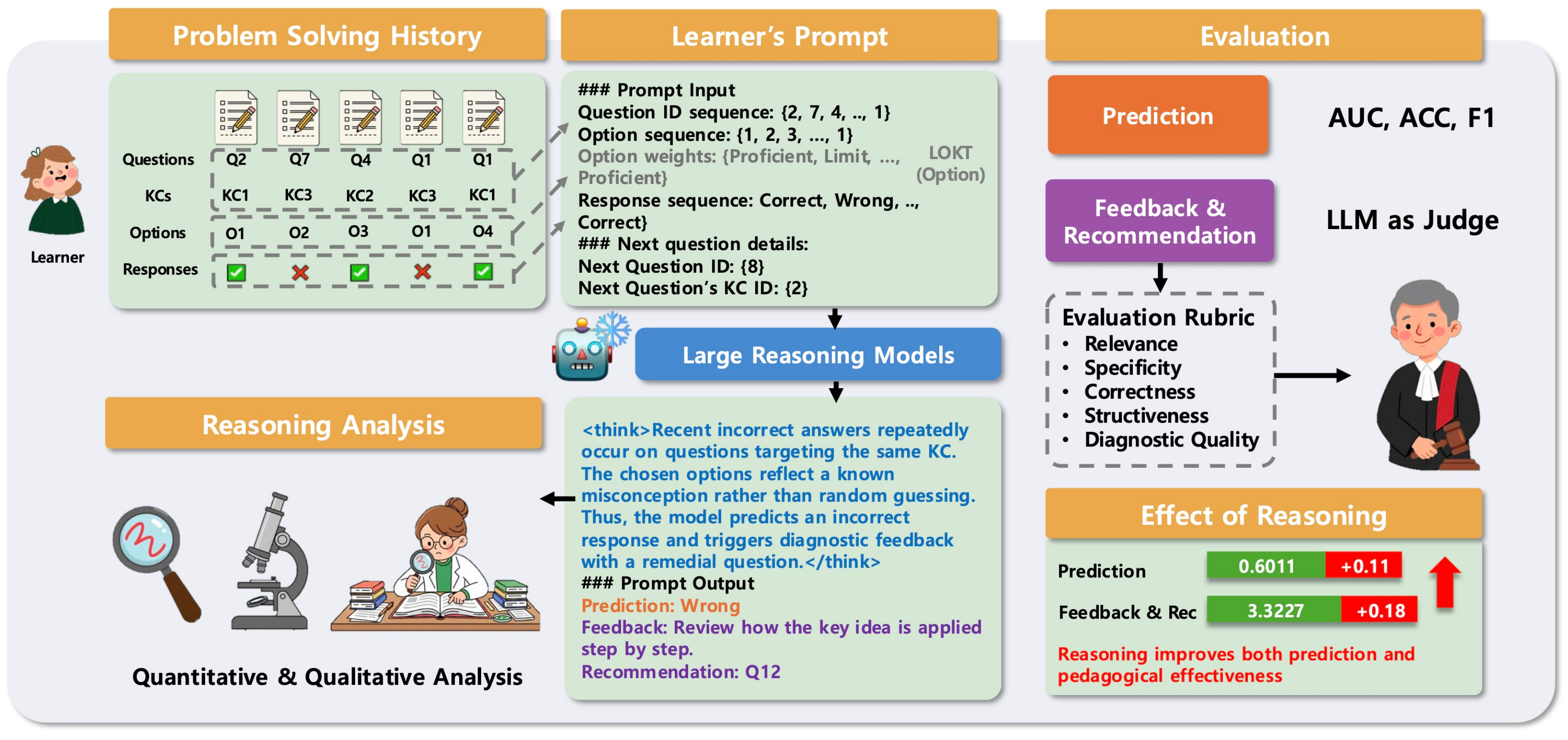}
    \caption{Overall framework of \textit{Thinking-KT}. Given a learner’s problem-solving history, the LLM receives a structured prompt and performs TTS before prediction. A single inference produces (1) KT prediction, (2) personalized diagnostic FB, and (3) next-step learning Rec. The \textit{right} panel illustrates the evaluation pipeline for both predictive performance and pedagogical quality, highlighting the positive effect of TTS on accuracy and instructional effectiveness.}
    \label{fig:main}
\end{figure*}

Knowledge Tracing (KT) aims to estimate a learner’s knowledge state over time based on their past interactions with learning tasks \cite{corbett1995knowledge}. Deep learning-based models \cite{piech2015deep, ghosh2020context, liu2023simplekt} have achieved strong predictive performance, yet fundamental limitations persist. Educational interaction data is typically too limited in scale for effective self-supervised pretraining, which leads to cold-start problems \cite{lee2024language, jung2024clst}. Moreover, existing KT models are primarily designed for prediction, requiring separate components to generate pedagogical feedback (FB) or learning recommendations (Rec) in modern intelligent tutoring systems (ITS), which incurs additional inference latency and system overhead \cite{vanlehn2011relative, ma2014intelligent}.

Recent LLM-based approaches have attempted to address these gaps \cite{fu2024sinkt, scarlatos2025dialoguekt, li2025explainable}, but without fine-tuning, they achieve only BKT-level performance \cite{cho2024systematic}, and even fine-tuned models fail to generalize. While LOKT \cite{kim2024lokt} demonstrated promising training-free KT through prompt engineering, its effectiveness was validated only on frontier Large Language Models (LLMs), and the role of Test-Time Scaling (TTS) was not examined.

The emergence of Large Reasoning Models (LRMs) such as OpenAI's o1 \cite{openai2024o1} and DeepSeek-R1 \cite{deepseek2025r1} has revealed that TTS is critical for complex tasks. Chain-of-Thought (CoT) prompting \cite{wei2022chain} and TTS \cite{snell2024scaling, xu2025lrmsurvey} reveal that explicit modeling of reasoning processes drives performance gains, rather than model size alone \cite{yang2025thinking}. However, no prior work has investigated how TTS influences KT performance, nor whether small LLMs can leverage TTS to become effective KT predictors.

To address these gaps, we propose \textit{Thinking-KT}, a training-free KT framework that performs TTS via structured reasoning and jointly generates KT predictions and pedagogical prescriptions, including FB and learning Rec, within a single inference. This framework enables small LLMs (1–2B parameters) to achieve competitive KT performance without additional training. Our experiments further show that TTS is the primary component of performance gains, while other inductive biases contribute only marginally. Moreover, by unifying prediction and prescription in a single inference, \textit{Thinking-KT} reduces reliance on complex multi-model ITS pipelines and provides interpretability through explicit reasoning traces.

Our contributions are summarized as follows:
\begin{itemize}
    \item We propose \textit{Thinking-KT}, a training-free KT framework that applies TTS via TTS and unifies prediction and pedagogical prescription.
    \item We show that TTS is the primary driver of performance gains in LLM-based KT.
    \item We reveal that a small LLM can jointly generate KT predictions, personalized FB, and learning Rec without degrading predictive accuracy.
    \item We present the systematic analysis of reasoning traces in KT, examining how the traces relate to both prediction accuracy and pedagogical quality.
\end{itemize}

%% file: 3_method.tex
\section{Method}

This section introduces \textit{Thinking-KT}, a training-free framework that incorporates TTS and enables a small LLM
to perform 
(1) KT prediction, 
(2) personalized FB generation, and
(3) learning Rec in a unified output.

\subsection{Problem Definition}

Let a learner's interaction history up to time $t$ be
\[
\mathcal{H}_t
=
\{(q_1, c_1, o_1, y_1), \ldots, (q_t, c_t, o_t, y_t)\},
\]
where $q_i$ denotes the question,
$c_i$ the associated knowledge component (KC),
$o_i$ the selected option,
and $y_i \in \{0,1\}$ the correctness label.

Given the next item $(q_{t+1}, c_{t+1})$,
the goal of KT is to estimate
\[
P(y_{t+1}=1 \mid \mathcal{H}_t, q_{t+1}, c_{t+1}).
\]

\textit{Thinking-KT} addresses this task using a structured natural-language prompt and a TTS mechanism, without any training or parameter updates.

\subsection{Thinking-KT}

\subsubsection{Structured Prompt Setting}

\textit{Thinking-KT} encodes a learner’s interaction history and the upcoming question into a structured natural-language prompt  (See Appendix \ref{appendix:structed_prompt_for_inference}).
To analyze the effect of prompt structure independently from reasoning, we consider three prompts: No-Option, Option and Weight.

In the No-Option, the prompt includes only the question and the learner’s past interaction history. The options are not provided. This setting represents a minimal-information baseline of LLM-based KT. In the Option, all answer options of the multiple-choice question (MCQ) are included in natural language. In the Weight, each answer option is augmented with a textual importance descriptor summarizing its diagnostic relevance. This representation follows prior work \cite{kim2024lokt} on textual option weighting for token-efficient LLM-based KT.

All three prompt settings are evaluated under both No-Think
and \textit{Thinking-KT} settings. This design allows us to disentangle the effect of prompt structure from the effect of TTS.

\subsubsection{TTS}

\textit{Thinking-KT} leverage TTS $\tau$ to generate at inference time. Given the structured prompt $\mathbf{x}$, the model performs TTS before producing its outputs. To study the effect of TTS, we impose a Thinking Budget $B$ that constrains the token length of the reasoning trace $\tau$ as:

\[
|\tau| \le B.
\]

The $B$ controls the depth of TTS and enables systematic analysis of how reasoning trace influences prediction accuracy and quality of FB and Rec.

\subsubsection{Unified Multi-Task Output}

A central component of \textit{Thinking-KT} is that a single LLM inference produces prediction and instructional outputs jointly. Specifically, the model generates:
\[
(\hat{y}_{t+1}, \hat{F}_{t+1}, \hat{R}_{t+1}, \tau),
\]
where $\hat{y}_{t+1}$ is the predicted probability of a correct response, $\hat{F}_{t+1}$ is FB describing the learner’s recent performance, $\hat{R}_{t+1}$ is a Rec for the next concept or practice item, and $\tau$ is the reasoning trace produced during TTS.

\subsubsection{Final Formulation}

We formally define the complete \textit{Thinking-KT} framework which is a unified, training-free TTS process. Given the structured prompt $\mathbf{x}$ and a Thinking Budget $B$,
a single LLM $f_{\text{LLM}}$ produces all outputs in one forward pass:
\[
(\hat{y}_{t+1}, \hat{F}_{t+1}, \hat{R}_{t+1}, \tau)
=
f_{\text{LLM}}(\mathbf{x}; B)
\]

where the reasoning trace $|\tau| \le B$.

% ==========================================================
% TABLE 1 — KT Prediction Performance
% ==========================================================
\begin{table*}[t]
\centering
\scriptsize
\caption{
KT prediction performance on ASSIST09, DBE-KT22, and EdNet. Within the \textit{Thinking-KT} framework, small LLMs outperform larger LLMs and achieve performance competitive with task-specific trained models, despite requiring no additional training.
}
\begin{tabular*}{\textwidth}{@{\extracolsep{\fill}}lccccccccccc}
\toprule
\multirow{2}{*}{\textbf{Model}} &
\multirow{2}{*}{\textbf{Design}} &
\multicolumn{3}{c}{\textbf{ASSIST09}} &
\multicolumn{3}{c}{\textbf{DBE-KT22}} &
\multicolumn{3}{c}{\textbf{EdNet-500}} \\
\cmidrule(lr){3-5} \cmidrule(lr){6-8} \cmidrule(lr){9-11}
 & & AUC & ACC & F1 & AUC & ACC & F1 & AUC & ACC & F1 \\
\midrule

%%%%%%%%%%%%%%%%%%%%%%%%%%%%%%%%%%%%%%%%%%%%%%%%%%%%%%%%%%%%%
\rowcolor{lightgray}
\multicolumn{11}{c}{\textbf{Traditional KT Models (Training)}} \\
\midrule
DKT      & --        & 0.7244 & - & - & 0.7820 & - & - & 0.7890* & - & - \\
DKVMN    & --        & 0.7199 & - & - & 0.7830 & - & - & 0.7800* & - & - \\
AKT      & --        & 0.7881 & - & - & 0.7980 & - & - & 0.8100* & - & - \\
\midrule

%%%%%%%%%%%%%%%%%%%%%%%%%%%%%%%%%%%%%%%%%%%%%%%%%%%%%%%%%%%%%
\rowcolor{lightgray}
\multicolumn{11}{c}{\textbf{LLM-based KT (No-Think / Training-Free)}} \\
\midrule

Qwen3-1.7B & No-Option     & 0.6011 & 0.7603 & 0.8540 & 0.5216 & 0.7253 & 0.8377 & 0.4930 & 0.5511 & 0.6716 \\
           & Option        & 0.5846 & 0.7493 & 0.8459 & 0.5262 & 0.7555 & 0.8581 & 0.5096 & 0.5783 & 0.7075 \\
           & Weight        & 0.5221 & 0.7621 & 0.8637 & 0.5039 & 0.7610 & 0.8643 & 0.5042 & 0.6012 & 0.7497 \\
\midrule

GPT-5-mini & No-Option     & 0.6437 & 0.4522 & 0.4710 & 0.5875 & 0.3471 & 0.2925 & 0.5117 & 0.4068 & 0.0142 \\
           & Option        & 0.6372 & 0.4404 & 0.4561 & 0.5818 & 0.3269 & 0.2414 & 0.6063 & 0.6132 & 0.6892 \\
           & Weight        & 0.6412 & 0.4494 & 0.4573 & 0.5750 & 0.2994 & 0.1747 & 0.5889 & 0.6016 & 0.6710 \\
\midrule

Gemini-2.5-flash & No-Option & 0.6992 & 0.7252 & 0.8126 & 0.7105 & 0.7527 & 0.8333 & 0.5875 & 0.5671 & 0.5781 \\
                 & Option    & 0.6991 & 0.7434 & 0.8272 & 0.6999 & 0.7445 & 0.8268 & 0.5884 & 0.5731 & 0.6048 \\
                 & Weight    & 0.6824 & 0.7322 & 0.8191 & 0.7105 & 0.7527 & 0.8333 & 0.6058 & 0.5511 & 0.5465 \\
\midrule

\rowcolor{lightblue}
\multicolumn{11}{c}{\textbf{\textit{Thinking-KT} (Thinking / Training-Free) (Ours)}} \\

\midrule

Qwen3-1.7B & No-Option     & 0.7276 & 0.7420 & 0.8271 & 0.6958 & 0.7390 & 0.8276 & 0.6046 & 0.5792 & 0.6209 \\
           & Option        & 0.7153 & 0.7630 & 0.8456 & 0.6597 & 0.6978 & 0.8057 & 0.6043 & 0.5932 & 0.6622 \\
           & Weight        & 0.7026 & 0.7511 & 0.8425 & 0.5953 & 0.7335 & 0.8336 & 0.5612 & 0.5531 & 0.6135 \\
\midrule

GPT-5-mini & No-Option   & 0.6614 & 0.7742 & 0.8576 & 0.6387 & 0.7719 & 0.8576 & 0.5992 & 0.5935 & 0.6666 \\
           & Option      & 0.6694 & 0.7841 & 0.8646 & 0.6367 & 0.7637 & 0.8522 & 0.6053 & 0.6124 & 0.6871 \\
           & Weight      & 0.6625 & 0.7666 & 0.8520 & 0.6327 & 0.7692 & 0.8536 & 0.6327 & 0.7692 & 0.8536 \\
\midrule

Gemini-2.5-flash & No-Option   & 0.6899 & 0.7252 & 0.8119 & 0.7122 & 0.7362 & 0.8208 & 0.6969 & 0.7388 & 0.8219 \\
                 & Option      & 0.7013 & 0.7406 & 0.8250 & 0.7185 & 0.7664 & 0.8446 & 0.5879 & 0.5671 & 0.6000 \\
                 & Weight      & 0.6824 & 0.7322 & 0.8191 & 0.6969 & 0.7388 & 0.8219 & 0.6004 & 0.5551 & 0.5524 \\

\bottomrule
\end{tabular*}
\label{tab:prediction}
\end{table*}

% Evaluation
\subsection{Evaluation}

\textit{Thinking-KT} produces KT prediction, FB and Rec. Each component is evaluated using appropriate metrics and protocols that remain consistent across all LLMs.

\subsubsection{Evaluation for KT prediction}

For KT prediction, we compute AUC, Accuracy, and F1 using a 10-sample evaluation protocol. Because each LLM forward pass yields only a binary correctness label (\texttt{correct} or \texttt{incorrect}), a single query does not provide probabilistic information required for AUC. Thus, each test instance is queried ten times, and the empirical correctness probability is used to compute AUC. Accuracy and F1 are computed via majority voting over the ten samples. Traditional KT models (DKT \cite{piech2015deep}, DKVMN \cite{zhang2017dynamic}, AKT \cite{ghosh2020context}) are deterministic and therefore do not require sampling.

\subsubsection{Evaluation for FB and Rec}

To evaluate the pedagogical quality of generated FB and Rec, we adopt an LLM-as-Judge \cite{zheng2023judging} using Solar-Pro2 \cite{upstage_console_solarpro2_docs_2025}
as the evaluator. Solar-Pro2 is selected as the judge model due to its favorable balance between performance and inference cost. For each test instance, a single generation is sampled from the model outputs and independently assessed by the judge model.

The evaluator assesses whether the generated FB and Rec appropriately reflect the learner’s interaction history, predicted mastery state, and pedagogical suitability, using a structured five-dimensional rubric: relevance (Rel), specificity (Spec), correctness (Corr), structiveness (Struct), and diagnostic quality (Diag), which rated on a 1–5 Likert scale (See Appendix \ref{appendix:prompt_feedback_evaluation}, \ref{appendix:prompt_recoomendation}). For joint FB and Rec outputs (FB+Rec), both rubric sets are applied. Scores are averaged across evaluation samples to produce the final generative quality metrics. The complete evaluation prompts and rubric definitions are provided in the Appendix.

%% file: 4_result.tex
% ==========================================================
% TABLE 2 — Unified Output Performance
% ==========================================================
\begin{table*}[t]
\centering
\scriptsize
\caption{
Unified output evaluation. Prediction accuracy is largely preserved when adding feedback and recommendation, and TTS further improves both prediction and pedagogical quality.
}
\begin{tabular*}{\textwidth}{@{\extracolsep{\fill}}l c c ccccc ccccc}
\toprule
\multirow{2}{*}{\textbf{Model}} &
\multirow{2}{*}{\textbf{Category}} &
\textbf{AUC} &
\multicolumn{5}{c}{\textbf{Feedback Rubrics}} &
\multicolumn{5}{c}{\textbf{Recommendation Rubrics}} \\
\cmidrule(lr){4-8} \cmidrule(lr){9-13}
& & 
& Rel & Spec & Corr & Struct & Diag
& Rel & Spec & Corr & Struct & Diag \\

 \midrule

\multirow{4}{*}{Qwen3-1.7B-No-Think}
 & Pred Only & 0.5221 & -- & -- & -- & -- & -- & -- & -- & -- & -- & -- \\
 & FB        & 0.7019    & 3.7683 & 2.8180 & 3.7951 & 3.1597 & 3.0726 & -- & -- & -- & -- & -- \\
 & Rec       & 0.6340    & -- & -- & -- & -- & -- & 3.2633 & 3.8883 & 3.4615 & 3.0543 & 2.2882 \\
 & FB + Rec  & 0.6857    & 3.8038 & 2.9874 & 3.7766 & 3.0974 & 3.2500 & 3.3480 & 3.9073 & 3.3652 & 3.1287 & 2.2876 \\

 \midrule

\multirow{4}{*}{Qwen3-1.7B-Think}
 & Pred Only & 0.7026 & -- & -- & -- & -- & -- & -- & -- & -- & -- & -- \\
 & FB        & 0.6810    & 3.8888 & 3.0037 & 3.9372 & 3.4354 & 3.5048 & -- & -- & -- & -- & -- \\
 & Rec       & 0.7183    & -- & -- & -- & -- & -- & 3.6424 & 4.0945 & 3.7241 & 3.4535 & 2.8960 \\
 & FB + Rec  & 0.7065    & 3.9342 & 3.2503 & 3.8893 & 3.3280 & 3.5445 & 3.4967 & 4.0140 & 3.5567 & 3.3422 & 2.5456 \\

\bottomrule
\end{tabular*}
\label{tab:unified_rubric}
\end{table*}

\section{Results}
\label{section:results}

This section reports experimental results of \textit{Thinking-KT}, focusing on the role of TTS in LLM-based KT and unified outputs.

\subsection{Experiment Settings}

All experiments are designed to isolate the effect of TTS from confounding factors such as prompt design and model scale. Our evaluation addresses two questions:  
(1) how TTS affects KT prediction performance and unified outputs, and  
(2) whether unified generation of prediction, feedback, and recommendation introduces a trade-off with KT accuracy.

We evaluate three categories of models: traditional KT architectures (DKT, DKVMN and AKT), small LLMs (Qwen3-1.7B), and frontier LLMs (GPT-5-mini and Gemini-2.5-Flash). All LLMs are evaluated under both reasoning-disabled and reasoning-enabled settings.

% dataset
% assistment09 (assist09), dbe-kt-22, ednet-500
Experiments are conducted on three benchmark KT datasets: ASSISTments09 (assist09), DBE-KT-22, and EdNet-500. ASSISTments09 is a middle-school mathematics tutoring dataset from the ASSISTments platform, comprising 4,217 students and 346,860 interactions over 26,688 problems with 123 knowledge components \cite{feng2009assistments}. DBE-KT-22 is a university-level database systems dataset containing 1,361 students and 167,222 interactions across 212 questions annotated with 98 skills \cite{abdelrahman2022dbekt22}. EdNet originates from a large-scale TOEIC preparation platform and includes 784,309 learners and 131M interactions in its full form \cite{choi2020ednet_aied}; following a low-data evaluation setting, we construct EdNet-500 by subsampling interaction histories from 500 learners. For all datasets, interaction sequences are randomly split into training and test sets with an 80/20 ratio.

In addition to prediction-only evaluation, we consider a unified output setting in which a single LLM jointly generates KT prediction, FB and Rec. Unified and prediction-only settings share identical inputs, Thinking Budgets, sampling protocols, and evaluation procedures; they differ only in output specification.

To analyze interactions between reasoning and prompt design, we compare three prompt formulations: No-Option, Option, and Weight. Unless otherwise stated, results use a history length of 25 interactions and a weight portion of 1, referencing \cite{kim2024lokt}.

\subsection{KT Performance}

Table~\ref{tab:prediction} reports KT prediction performance across datasets and model categories. Without TTS, LLM-based KT shows unstable behavior across prompt formulations. Performance varies substantially depending on option representations, and in several cases degrades when additional option information is introduced, indicating that prompt-level inductive bias alone is insufficient.

When TTS is enabled, KT performance improves consistently across datasets and prompt designs. As shown in Table~\ref{tab:prediction}, performance becomes substantially more robust to prompt variations: while No-Option, Option, and Weight prompts yield divergent results without TTS, their performance differences narrow under reasoning-enabled settings. This indicates that TTS dominates prompt-level effects.

The impact of TTS is particularly pronounced for small open-source models. For Qwen3-1.7B, reasoning yields large AUC gains across datasets, transforming unstable predictions into competitive KT performance and narrowing the gap with larger frontier LLMs. Overall, Table~\ref{tab:prediction} shows that TTS plays a more critical role than prompt complexity or model scale in stabilizing KT predictions and improving ranking quality.

\subsection{Unified Output Performance}

We examine whether unified generation introduces a trade-off with prediction accuracy. Table~\ref{tab:unified_rubric} summarizes results across four output configurations.

Unified generation does not degrade KT prediction performance: AUC remains comparable between Pred Only and unified settings. This indicates that generating pedagogical outputs does not inherently interfere with knowledge state estimation.

TTS plays a key role in stabilizing unified generation. For Qwen3-1.7B, KT performance varies across output configurations when reasoning is disabled, suggesting interference between prediction and pedagogical generation. With TTS enabled, AUC remains stable across Pred Only, FB, Rec, and FB+Rec settings, indicating that reasoning enables joint generation without performance trade-offs.

In addition to preserving KT accuracy, TTS improves pedagogical quality. As shown in Table~\ref{tab:unified_rubric}, \textit{Thinking-KT} achieve higher scores across Rel, Spec, Corr, Struct and Diag. Overall, these results demonstrate that TTS enables unified outputs without sacrificing KT performance, allowing a single LLM to function as a unified ITS engine.

%%%%%%%%%%%%%%%%%%%%%%%%%%%%%%%%%%%%%%%%%%%%%%%%%
%%%%%%5 analysis에 속하는 figure (위치문제로 여기에 배치)
%%%%%%%%%%%%%%%%%%%%%%%%%%%%%%%%%%%%%%%%%%%%%%%%%
\begin{figure*}[t]
\centering
\hspace*{-3em}
\captionsetup[subfigure]{skip=-0.1em}
\begin{subfigure}[b]{0.48\textwidth}
    \centering
    \includegraphics[width=\linewidth]{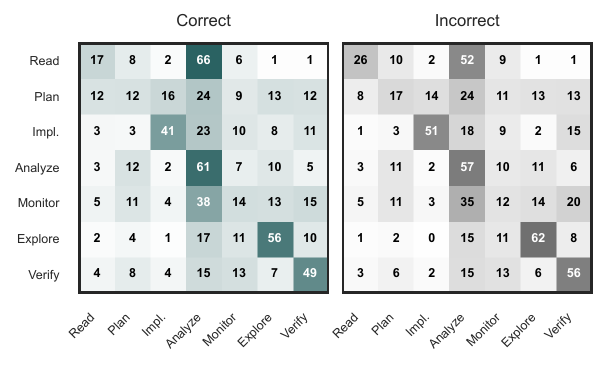}
    \caption{
    Label transition patterns
    }
    \label{fig:analysis_a}
\end{subfigure}
\begin{subfigure}[b]{0.271\textwidth}
    \centering
    \includegraphics[width=\linewidth]{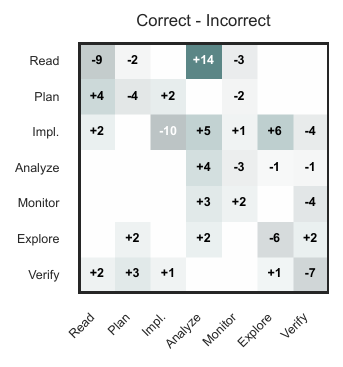}
    \caption{
    Transition difference
    }
    \label{fig:analysis_b}
\end{subfigure}
\begin{subfigure}[b]{0.27\textwidth}
    \centering
    \raisebox{0.8em}{%
    \includegraphics[width=\linewidth]{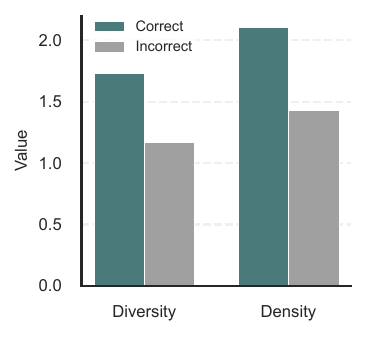}
    }
    \caption{
    Trace complexity
    }
    \label{fig:analysis_c}
\end{subfigure}
\vspace{-0.5em}
\caption{
Structural analysis of reasoning traces.
Correct predictions are characterized by analytically centered transition structures,
greater diversity, and denser transition patterns, whereas incorrect predictions exhibit
more fragmented and self-referential dynamics.
}
\label{fig:analysis}
\end{figure*}

%% file: 5_ablation.tex
%%%%%%%%%%%%%%%%%%%%%%%%%%%%%%%%%%%%%%%%%%%%%%%%%%%%%%%%%
% Ablation
%%%%%%%%%%%%%%%%%%%%%%%%%%%%%%%%%%%%%%%%%%%%%%%%%%%%%%%%%

\section{Ablation}

% Thinking budget에 따른 성능 변화
\subsection{Comparison with Thinking Budget}

Table~\ref{tab:thinking_budget} shows the impact of the Thinking Budget on KT performance. Introducing TTS yields a substantial AUC improvement over the No-Think setting (0.5221 → 0.7026 at 2048 tokens), while further increasing the Thinking Budget to 4096 tokens does not provide additional gains and slightly degrades performance. This indicates that KT benefits primarily from the presence of TTS rather than excessive Thinking Budget, suggesting an optimal and task-specific range are needed.

\begin{table}[t]
\centering
\small
\caption{
Effect of Thinking Budget on KT prediction performance for Qwen3-1.7B.
}
\label{tab:thinking_budget}
\begin{tabular}{lccc}
\toprule
\textbf{Thinking Budget} & \textbf{AUC} & \textbf{ACC} & \textbf{F1} \\
\midrule
No-Think   & 0.5221 & 0.7621 & 0.8637 \\
Think-2048    & \textbf{0.7026} & 0.7511 & 0.8425 \\
Think-4096    & 0.6863 & 0.7411 & 0.8360 \\
\bottomrule
\end{tabular}
\end{table}

% history len, fewshot for weight (portion)에 대한 성능 변화 확인
\subsection{Importance of History Length and Few-shot Prompt Structure}

\begin{figure}[t]
    \centering
    \vspace{-1em}
    \includegraphics[width=\columnwidth]{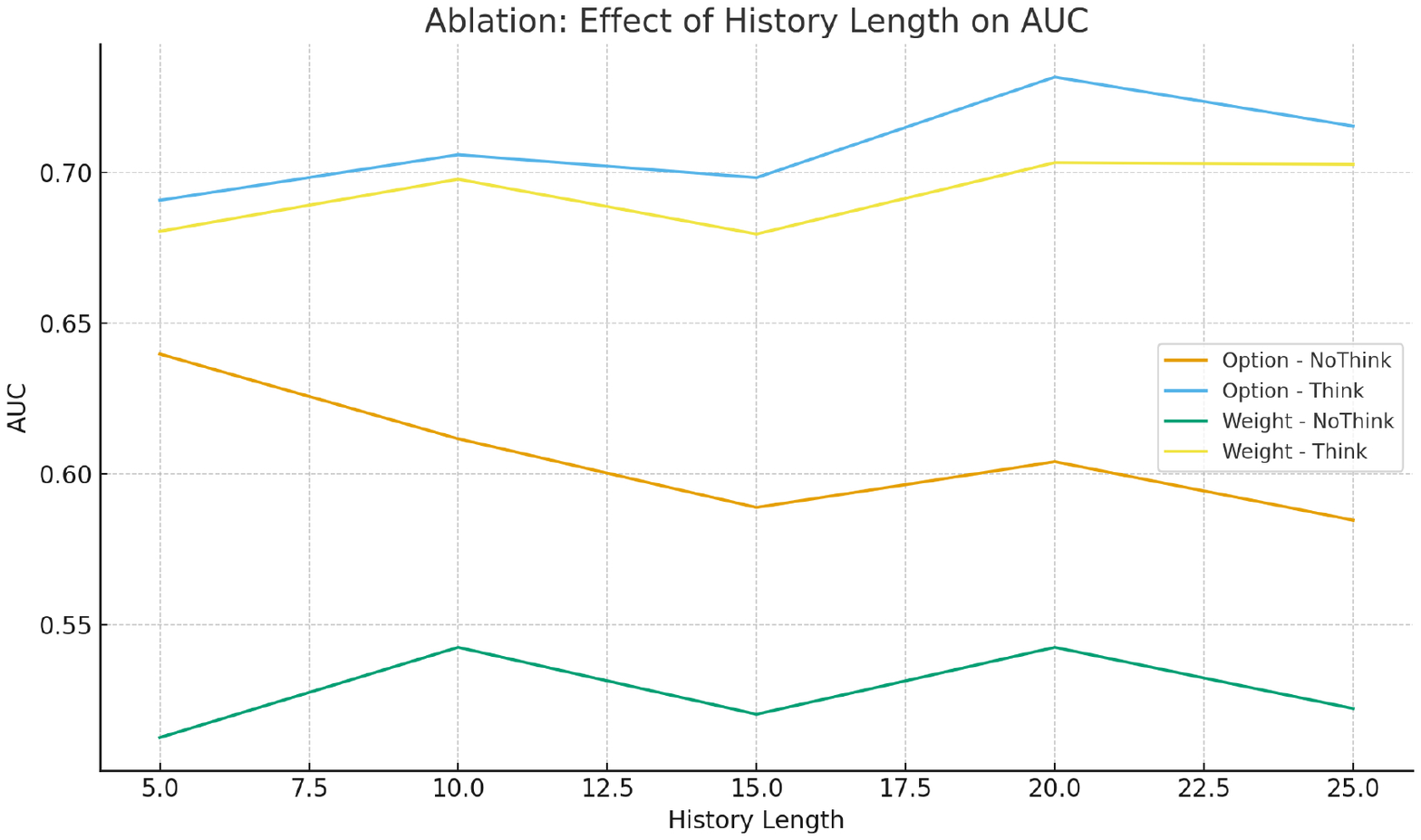}
    \vspace{-2.5em}
    \caption{
    Effect of history length on AUC. Thinking-KT (2048 tokens) consistently improves performance across all lengths, with saturation around 20--25 interactions.
    }
    \label{fig:history}
\end{figure}

Figure~\ref{fig:history} shows that enlarging the history window alone provides limited benefits without TTS, whereas \textit{Thinking-KT} yields substantial AUC gains (+0.10–0.17) across prompt settings, with performance saturating at moderate history lengths. This suggests that TTS, rather than prompt complexity, is the primary factor driving KT improvements.

\subsection{Effect of Model Size}

We analyze the effect of model scale using three Qwen3 variants (0.6B, 1.7B, and 4B) under identical settings on ASSIST09, with the same structured prompt (Weight) and a fixed reasoning budget of 2048 tokens. As shown in Table~\ref{tab:model_size}, TTS yields meaningful KT performance even for the smallest model (AUC = 0.6168), while scaling to 1.7B achieves the best AUC (0.7276); further scaling to 4B does not improve AUC and slightly degrades it. In contrast, Accuracy and F1 increase monotonically with model size, suggesting that larger models make more confident binary decisions, whereas the ranking quality of knowledge state estimation saturates at moderate scales, indicating that TTS rather than model size is the primary driver of KT performance.

\begin{table}[t]
\centering
\small
\caption{
Effect of model size on KT prediction performance under \textit{Thinking-KT}
(Thinking Budget = 2048).
}
\label{tab:model_size}
\begin{tabular}{lccc}
\toprule
\textbf{Model Size} & \textbf{AUC} & \textbf{ACC} & \textbf{F1} \\
\midrule
Qwen3-0.6B & 0.6168 & 0.7291 & 0.8308 \\
Qwen3-1.7B & \textbf{0.7276} & 0.7420 & 0.8271 \\
Qwen3-4B   & 0.7182 & \textbf{0.7775} & \textbf{0.8579} \\
\bottomrule
\end{tabular}
\end{table}

%% file: 6_analysis.tex
\section{Analysis}
This section examines the reasoning traces of \textit{Thinking-KT}. 
We first define an analytical framework based on Schoenfeld's Episode Theory, 
then examine structural patterns distinguishing correct from incorrect predictions, 
their pedagogical interpretation, and the qualitative impact on feedback and recommendations.

%%%%%%%%%%%%%%%%%%%%%%%%%%%%%%%%%%%%%%%%%%%%
%%%%%%%%%%%%%%%%%%%%%%%%%%%%%%%%%%%%%%%%%%%%
\subsection{Experimental Setup}
To isolate the impact of TTS, we compare two settings applied to Qwen3-1.7B model. The No-Think baseline requires the model to predict the final KT outcome directly from the input prompt, without generating explicit reasoning traces. In contrast, the \textit{Thinking-KT} framework allocates a reasoning budget of 2048 tokens, allowing the model to externalize its cognitive process as a reasoning trace before outputting the final answer. Our framework adapts Schoenfeld's Episode Theory ~\cite{schoenfeld1985mathematical} to categorize 11,082 reasoning traces. As shown in Table~\ref{tab:taxonomy}, each trace is mapped to seven cognitive labels. The labels are designed to assess the depth and structural integrity of the TTS process. Following~\ref{section:results}, we use Solar-Pro2~\cite{upstage_console_solarpro2_docs_2025} for labeling. 

%%%%%%%%%%%%%%%%%%%%%%%%%%%%%%%%%%%%%%%%%%%%
%%%%%%%%%%%%%%%%%%%%%%%%%%%%%%%%%%%%%%%%%%%%
\begin{table}[t]
\centering
\small
\caption{Taxonomy of reasoning traces based on Schoenfeld's Episode Theory.}
\label{tab:taxonomy}
\vspace{-3pt}
\begin{tabular}{l p{0.65\linewidth}}
\toprule
Label & Description \\
\midrule
\texttt{Read}      & Identifying problem and student history.\\
\texttt{Plan}      & Formulating pedagogical strategies and subsequent steps. \\
\texttt{Implement} & Executing calculations or logical operations. \\
\texttt{Analyze}   & Evaluating theoretical relationships and mastery states. \\
\texttt{Monitor}   & Regulating and calibrating the reasoning flow in real-time. \\
\texttt{Explore}   & Testing hypotheses through trial-and-error reasoning.\\
\texttt{Verify}    & Confirming the consistency and correctness of results. \\
\bottomrule
\end{tabular}
\end{table}

\begin{table}[h]
\centering
\scriptsize
\resizebox{\columnwidth}{!}{
\begin{tabular}{lccc}
\toprule
\textbf{Pattern} & \textbf{Think} & \textbf{No-Think} & \textbf{$\Delta$} \\
\midrule
Evidence specificity & 78\% & 24\% & +54pp \\
Performance differentiation & 64\% & 18\% & +46pp \\
Prediction-feedback alignment & 86\% & 62\% & +24pp \\
\bottomrule
\end{tabular}}
\caption{Feedback pattern frequencies across conditions ($n$=50 per condition).}
\label{tab:feedback_patterns}
\end{table}

\subsection{Structural Systematicity of Reasoning Traces}

Analysis of reasoning traces reveals three major differences of structural systeaticity between correct and incorrect predictions.

First, we observe that correct predictions tend to show rapid transition from problem comprehension to analysis. As shown in Figure~\ref{fig:analysis_a}, 
66.1\% of \texttt{Read} episodes transition to \texttt{Analyze} in correct 
predictions, compared to 51.7\% in incorrect predictions. This pattern is consistent with the hypothesis that effective KT may benefit from immediate engagement with the underlying problem structure through \texttt{Read}$\rightarrow$\texttt{Analyze}, though causal relationships cannot be established from this observational analysis.

Second, incorrect predictions exhibit undirected self-referential patterns, 
as highlighted by the red regions in Figure~\ref{fig:analysis_b}. 
\texttt{Explore}$\rightarrow$\texttt{Explore} transitions are 6.2pp higher 
in incorrect traces, \texttt{Verify}$\rightarrow$\texttt{Verify} loops are 
7.1pp higher, and notably, \texttt{Implement}$\rightarrow$\texttt{Implement} 
shows a 10.2pp gap. These patterns indicate that incorrect predictions 
become trapped in repetitive cognitive cycles without analytical guidance, 
cycling through exploration or verification without reaching productive 
conclusions.

Third, beyond transition dynamics, correct predictions demonstrate 
systematically higher trace complexity (Figure~\ref{fig:analysis_c}). 
This suggests an optimal range of reasoning depth for knowledge tracing tasks.

These findings suggest that \textit{Thinking-KT}'s performance gains and robustness are associated with structured analytical reasoning patterns. 

\begin{table}[t]
\vspace{-1em}
\centering
\scriptsize
\resizebox{\columnwidth}{!}{
\begin{tabular}{lll}
\toprule
\textbf{Episode Label} & \textbf{Diagnostic Analogue} & \textbf{Evidence} \\
\midrule
\texttt{Read} & Evidence gathering & Initiates 89\% of traces \\
\texttt{Read}$\rightarrow$\texttt{Analyze} & Cue interpretation & 66\% (correct) vs 51\% (incorrect) \\
\texttt{Analyze} & Judgment formation & 40.83\% vs 34.80\% centrality \\
\bottomrule
\end{tabular}
}
\caption{Mapping between Schoenfeld's episode labels and teacher diagnostic competence components.}
\label{tab:diagnostic_mapping}
\end{table}

% Section 5.4: Pedagogical Interpretation of Thinking
% To be inserted after Section 5.3 in the main paper

\subsection{Pedagogical Interpretation of Thinking}

We interpret reasoning traces through the lens of \textit{teacher diagnostic competence} which is the cognitive processes of teachers who employ to assess student knowledge states and inform instructional decisions~\cite{helmke1987interactional,sudkamp2012accuracy}.

The reasoning patterns observed in \textit{Thinking-KT} exhibit structural analogies to expert diagnostic reasoning. Table~\ref{tab:diagnostic_mapping} maps Schoenfeld's episode labels to diagnostic competence components.

This mapping aligns with~\citet{herppich2018teachers}'s three-component model: (1) perceiving relevant student cues, (2) interpreting cues against domain knowledge, and (3) integrating information into judgments. The importance of \texttt{Analyze} in correct predictions corresponds to the cue interpretation phase, which~\citet{sudkamp2012accuracy} identified as the primary differentiator between accurate and inaccurate teacher judgments ($k$=75 studies). Conversely, incorrect predictions show elevated \texttt{Explore} and fragmented \texttt{Verify} patterns, analogous to novice diagnostic behavior characterized by insufficient analytical depth.

These findings indicate that TTS organizes the model’s assessment process into a diagnostic reasoning traces that resembles how expert teachers reason. However, this similarity should be understood as structural rather than functional.

% Section 5.5: Qualitative Analysis of Feedback and Recommendation

\subsection{Qualitative Analysis of Feedback and Recommendation}
\label{sec:qualitative_analysis}

We conducted exploratory qualitative analysis based on ground theory~\cite{glaser1965constant} to characterize how TTS affects the pedagogical properties of generated outputs. A single researcher performed open coding on 200 samples (50 per condition: Think-FB, No-Think-FB, Think-Rec, No-Think-Rec), generating 16 codes that were grouped into 5 emergent patterns through constant comparison.

Three patterns distinguished Think from No-Think feedback (Table~\ref{tab:feedback_patterns}): \textit{Evidence Specificity} (citing question IDs vs. general KC references), \textit{Performance Differentiation} (distinguishing success/struggle areas vs. uniform characterization), and \textit{Prediction-Feedback Alignment} (internal consistency). These patterns relate to~\citet{hattie2007power}'s feedback model, where effective feedback addresses ``How am I going?'' with concrete instances and enables targeted ``Where to next?'' guidance.

Two patterns emerged for recommendations: \textit{Gap-Targeting} (recommending struggled items; Think 72\% vs. No-Think 34\%) and \textit{Rationale Provision} (explaining why; Think 68\% vs. No-Think 22\%). The Gap-Targeting pattern reflects~\citet{narciss2008feedback}'s distinction between corrective and confirmatory strategies, with Think exhibiting predominantly corrective orientation.

%% file: 2_related_work.tex
\section{Related Work}

KT aims to predict students' knowledge states and future performance. Bayesian Knowledge Tracing (BKT) \cite{corbett1995knowledge} pioneered this field, and DKT \cite{piech2015deep} revolutionized it by applying deep learning. Subsequent research has advanced through attention \cite{ghosh2020context, liu2023simplekt, yin2023tracing}, graph learning \cite{cheng2024dygkt}, and uncertainty-aware modeling \cite{cheng2025uncertainty, xia2025flatformer}. However, KT research faces persistent challenges: the trade-off between performance and interpretability \cite{bai2024explainable, sun2024interpretable}, and the cold-start problem \cite{jung2024clst}, as educational data lacks the scale amenable to self-supervised learning \cite{lee2024language}. 

Recent LLM-based approaches have attempted to address these limitations \cite{fu2024sinkt, scarlatos2025dialoguekt, li2025explainable}. However LLMs achieve only BKT-level performance \cite{cho2024systematic}, and even fine-tuned models often fail to generalize beyond specific scenarios \cite{jung2024clst, lee2024language, kim2024lokt}. While LOKT \cite{kim2024lokt} showed promise through efficient textual encoding, critical gaps remain: reasoning capabilities and smaller models remain largely untested.

In parallel, LRMs have demonstrated that explicit multi-step reasoning, rather than model size alone, can drive substantial performance gains across diverse domains \cite{wei2022chain, snell2024scaling, xu2025lrmsurvey}. However, despite these advances, the role of TTS remains largely unexplored in educational settings \cite{lee2025pedagogy}, particularly in KT, motivating our proposal of \textit{Thinking-KT}.

%% file: 7_conclusion.tex
\section{Conclusion}

We proposed \textit{Thinking-KT}, a training-free framework for LLM-based KT that leverages reasoning to enable prediction and unified pedagogical generation. Experimental results show that reasoning is the dominant factor for stabilizing KT performance, surpassing the effects of prompt design and model scale, and allows small LLMs to achieve competitive performance without task-specific training. We further demonstrate that TTS enables a single LLM to jointly generate KT predictions, FB, and Rec without degrading prediction accuracy, eliminating the need for multi-stage ITS pipelines. These findings highlight TTS as a practical foundation for building lightweight, unified, and scalable educational AI systems.

\section{Limitation}

Despite its effectiveness, \textit{Thinking-KT} has several limitations that suggest directions for future research. 
First, although the framework is training-free, test-time Thinking introduces additional inference cost and latency, which may restrict its deployment in strict real-time intelligent tutoring systems. Future work will explore adaptive or early-exit reasoning mechanisms that dynamically allocate Thinking Budget based on uncertainty or task difficulty, reducing unnecessary computation. Second, while the observed reasoning traces exhibit structural similarities to expert teacher diagnostic processes, they should not be interpreted as faithful representations of human cognition. Future studies will incorporate human-in-the-loop evaluations, such as expert teacher annotations or think-aloud protocols, to more rigorously assess the pedagogical validity and interpretability of model reasoning. Third, the pedagogical quality of feedback and recommendations is evaluated using an LLM-as-Judge with structured rubrics, which may still introduce implicit model bias. Future work will complement automated evaluation with human expert assessment and investigate cross-model or ensemble judging strategies to improve robustness. Fourth, our experiments are limited to multiple-choice knowledge tracing benchmarks. Generalization to open-ended problem solving, dialog-based tutoring, and multi-turn instructional interactions remains an open challenge. Extending \textit{Thinking-KT} to these settings will be a key direction for future research. Finally, we observe that the optimal Thinking Budget is task- and dataset-dependent, indicating that a fixed reasoning depth may be suboptimal. Future work will investigate adaptive reasoning control policies that adjust Thinking Budget in response to learner state, task complexity, or confidence signals, enabling more efficient and personalized KT.

\section{GenAI Usage Disclosure}

GPT-5, 5.1 and 5.2 were used for minor language editing.  
Gemini-2.5-Flash, GPT-5-Mini, and Solar-Pro2 were used solely to run model predictions and evaluations. All scientific ideas, analyses, and experimental designs were conducted by the human authors.

%% file: 8_appendix.tex
\section{Appendix}
\label{sec:appendix}

\subsection{prompt: structured prompt for inference}
\label{appendix:structed_prompt_for_inference}

\begin{promptbox}
Analyze the student's problem-solving history and predict their performance on the next question.

Student's problem-solving history:
1. Question ID sequence: \{question\_ids\}
2. KC ID sequence: \{kc\_ids\} (Note: This is a list of KC IDs associated with the next question. For example, if next\_kc\_id is \{3, 72\}, it means the next question involves KC IDs 3 and 72.)
3. Selected option sequence: \{option\_sequence\}
4. Selected option weights: \{option\_weights\}
5. Correctness sequence: \{answer\_sequence\}

Next question details:
1. Next question ID: \{next\_question\_id\}
2. Next question's KC ID: \{next\_kc\_id\}

Based on the above information, predict whether the student will answer the next question
(ID: \{next\_question\_id\}, KC ID: \{next\_kc\_id\}) correctly (`correct`) or incorrectly (`wrong`).

Consider the following when making your prediction:
1. The student's overall correctness pattern.
2. The complexity and difficulty levels of the questions and KC IDs.
3. How the selected options reflect their weight on the student's understanding and confidence.
4. Recent trends in the student's performance.
5. The student's progression and knowledge improvement over time.
6. The student's current knowledge state for each KC and how it matches the KC IDs in the next question.
7. Ignore any NaN values in the option weights.

Output only the single word [`correct` or `wrong`]. No other words or punctuation should be included.

\end{promptbox}

\subsection{prompt: feedback evaluation}
\label{appendix:prompt_feedback_evaluation}

\begin{promptbox}

You are an expert educational assessment evaluator. Your task is to evaluate the quality of personalized feedback generated by a knowledge tracing system.

Context:
\begin{itemize}
    \item Student History: <STUDENT\_HISTORY>
    \item Prediction: <PREDICTION>
    \item Ground Truth: <GROUND\_TRUTH>
    \item Generated Feedback: <GENERATED\_FEEDBACK>
\end{itemize}

Evaluation Task:
Evaluate the generated feedback using the rubric below. For each criterion, assign a score from 1 to 5 and provide a brief explanation.

Rubric:

1. Relevance (1–5):
Does the feedback directly address the student’s learning history and current knowledge state?
1 = Completely irrelevant or generic
2 = Somewhat related but mostly generic
3 = Moderately relevant with limited personalization
4 = Highly relevant with clear personalization
5 = Perfectly tailored to the student’s specific situation

2. Specificity (1–5):
Does the feedback provide concrete, actionable guidance?
1 = Extremely vague, no actionable guidance
2 = Mostly generic statements
3 = Some specific elements mixed with generic advice
4 = Mostly specific with clear guidance
5 = Highly specific with detailed, actionable steps

3. Accuracy (1–5):
Is the feedback consistent with the student’s performance history and prediction outcome?
1 = Contradicts evidence
2 = Partially inconsistent
3 = Generally consistent with minor inaccuracies
4 = Accurate and consistent
5 = Perfectly aligned with all evidence

4. Constructiveness (1–5):
Does the feedback support learning improvement or reinforcement?
1 = No constructive value
2 = Minimal constructive guidance
3 = Some helpful elements
4 = Clear and constructive guidance
5 = Highly motivating and instructional feedback

5. Diagnostic Quality (1–5):
Does the feedback identify specific knowledge gaps or strengths based on KC patterns?
1 = No diagnostic insight
2 = Vague or incorrect diagnosis
3 = Basic diagnostic insight
4 = Clear identification of gaps or strengths
5 = Deep diagnostic insight with clear pedagogical implications

Output Format:
Return a JSON object with a score and explanation for each criterion.

\end{promptbox}

\subsection{prompt: recommendation evaluation}
\label{appendix:prompt_recoomendation}

\begin{promptbox}

You are an expert educational assessment evaluator. Your task is to evaluate the quality of question recommendations generated by a knowledge tracing system.

Context:
\begin{itemize}
    \item Student History: <STUDENT\_HISTORY>
    \item Prediction: <PREDICTION>
    \item Ground Truth: <GROUND\_TRUTH>
    \item Generated Feedback: <GENERATED\_FEEDBACK>
\end{itemize}

Evaluation Task:
Evaluate the generated recommendation using the rubric below. For each criterion, assign a score from 1 to 5 and provide a brief explanation.

Rubric:

1. Relevance (1–5):
Does the recommendation address the student’s learning history and current knowledge state?
1 = Completely irrelevant
2 = Weakly related, mostly generic
3 = Moderately relevant
4 = Highly relevant and personalized
5 = Perfectly tailored to the student’s knowledge gaps

2. Specificity (1–5):
Does the recommendation specify a concrete question and clear reasoning?
1 = No specific question or vague reasoning
2 = Question present but generic reasoning
3 = Question with some specific reasoning
4 = Specific question with clear reasoning
5 = Highly specific question with detailed pedagogical rationale

3. Accuracy (1–5):
Is the recommended question grounded in the student’s interaction history and evidence?
1 = Question not in history or reasoning incorrect
2 = Question in history but reasoning partially incorrect
3 = Generally accurate with minor issues
4 = Accurate and consistent
5 = Perfectly aligned with all evidence

4. Constructiveness (1–5):
Does the recommendation support learning improvement or reinforcement?
1 = No constructive value
2 = Minimal learning value
3 = Some constructive elements
4 = Clear pedagogical benefit
5 = Strongly supports learning progress

5. Diagnostic Quality (1–5):
Does the recommendation reflect specific knowledge gaps or strengths based on KC patterns?
1 = No diagnostic basis
2 = Vague or incorrect diagnosis
3 = Basic diagnostic insight
4 = Clear identification of relevant gaps or strengths
5 = Deep diagnostic analysis linked to learning goals

Output Format:
Return a JSON object with a score and explanation for each criterion.

\end{promptbox}

\subsection{prompt: labeling reasoning traces}
\label{appendix:trace_labeling_prompt}

We use Solar-Pro2~\cite{upstage_console_solarpro2_docs_2025} to classify reasoning traces based on Schoen-
feld’s Episode Theory~\cite{schoenfeld1985mathematical}.The following prompt is used for labeling:

\begin{promptbox}
You are an expert in analyzing mathematical problem-solving processes based on Schoenfeld's Episode Theory.

Given a segment of reasoning trace from a knowledge tracing model, classify it into one of the following categories:

1. \texttt{Read}: Identifying problem and student history. (e.g., ``The problem asks to find the value of $x$ when $2x+5=10$.'')

2. \texttt{Plan}: Formulating pedagogical strategies and subsequent steps. (e.g., ``Next, I will differentiate both sides of the equation with respect to $x$.'')

3. \texttt{Implement}: Executing calculations or logical operations. (e.g., ``Substituting $x=3$ into the equation: $2(3)+5=6+5=11$.'')

4. \texttt{Analyze}: Evaluating theoretical relationships and mastery states. (e.g., ``According to the Pythagorean theorem, in a right triangle, the square of the hypotenuse equals the sum of squares of the other two sides.'')

5. \texttt{Monitor}: Regulating and calibrating the reasoning flow in real-time. (e.g., ``Wait, something seems off here.'')

6. \texttt{Explore}: Testing hypotheses through trial-and-error reasoning. (e.g., ``Perhaps substituting different values for $x$ might reveal a pattern.'')

7. \texttt{Verify}: Confirming the consistency and correctness of results. (e.g., ``Let me check my calculation again... Yes, it matches the previous result.'')

Reasoning Trace Segment: <SEGMENT>

Output: Return only the label name (\texttt{Read}, \texttt{Plan}, \texttt{Implement}, \texttt{Analyze}, \texttt{Monitor}, \texttt{Explore}, or \texttt{Verify}).
\end{promptbox}